\documentclass[letterpaper, 10 pt, conference]{ieeeconf}  

\usepackage{amsmath}
\usepackage{moreverb,url}
\usepackage[]{hyperref}
\usepackage{algorithm}
\usepackage{algpseudocode}
\usepackage{dirtytalk}
\usepackage{graphicx} 
\usepackage{subfigure}
\usepackage{wrapfig}
\usepackage{url}
\usepackage{textcase}

\IEEEoverridecommandlockouts                              

\title{\LARGE \bf
A Bayesian-Based Approach to Human Operator Intent Recognition in Remote Mobile Robot Navigation}

\author{Dimitris Panagopoulos$^{1}$, Giannis Petousakis$^{1}$, Rustam Stolkin$^{1}$, Grigoris Nikolaou$^{2}$, Manolis Chiou$^{1}$
\thanks{}
\thanks{This work was supported by the UKRI-EPSRC grand EP/R02572X/1 (National Centre for Nuclear Robotics).}%
\thanks{$^{1}$Extreme Robotics Lab and National Centre for Nuclear Robotics, University of Birmingham, UK,
        {\tt\small \{d.panagopoulos, i.petousakis, r.stolkin, m.chiou\}@bham.ac.uk}}%
\thanks{$^{2}$University of West Attica, Greece
        {\tt\small \{nikolaou@uniwa.gr\}}}%

}

\begin{document}
\setlength{\parskip}{0em}
\bstctlcite{IEEEexample:BSTcontrol}

\maketitle
\thispagestyle{empty}
\pagestyle{empty}

\begin{abstract}
This paper addresses the problem of human operator intent recognition during teleoperated robot navigation. In this context, recognition of the operator's intended navigational goal, could enable an artificial intelligence (AI) agent to assist the operator in an advanced human-robot interaction framework. We propose a Bayesian Operator Intent Recognition (BOIR) probabilistic method that utilizes: (i) an \textit{observation} model that fuses information as a weighting combination of multiple observation sources providing geometric information; (ii) a \textit{transition} model that indicates the evolution of the state; and (iii) an \textit{action} model, the Active Intent Recognition Model (AIRM), that enables the operator to communicate their explicit intent asynchronously. The proposed method is evaluated in an experiment where operators controlling a remote mobile robot are tasked with navigation and exploration under various scenarios with different map and obstacle layouts. Results demonstrate that BOIR outperforms two related methods from literature in terms of accuracy and uncertainty of the intent recognition.
\end{abstract}


\section{Introduction}
When dealing with human teams, e.g. in sports or other cooperative tasks, the communication and inference of each team member's intent greatly affects the team's overall performance. Each member is involved in inferring the intention of others, and uses this intention to plan their own actions. Upon doing so, everyone performs in a complementary way towards a common goal, making the interaction efficient \cite{Goodrich2003}. Similarly to human teams, robotic technologies increasingly demand collaboration and interaction between humans (e.g. robot operators) and robots (e.g. embodied AI agents) to carry out tasks as a team. Unlike human beings that naturally perceive and communicate other people's intent, robots struggle with recognizing human intent or leveraging context out of human actions. Human-robot teams, just as in traditional human cooperation tasks, require the ability to discern each party's intent, cooperate and act as one in response to changes in the environment. Therefore, introducing human intent recognition capabilities to robotic systems is essential towards achieving optimal team performance \cite{Liu2016}.

This paper focuses on remotely operated mobile robots in high-consequence and safety-critical applications such as disaster response, remote inspection, nuclear decommissioning and telepresence. However, we propose ideas which may be applicable to other tasks and types of robot. To engage in these cooperative activities with and actively assist human operators, such robotic systems are progressing towards been equipped with various autonomous capabilities. To this end, the robot's AI agent needs to be aware of what the operator (i.e. the human agent) is trying to accomplish or what their goal is. This will enable the AI agent to inform and better adapt its policies based on the estimated operator's goal for efficient use of its task assisting capabilities. This is especially true for variable autonomy robotic systems such as shared control \cite{Pappas2020} or Mixed-Initiative (MI) \cite{Jiang2015} systems which can greatly benefit from the operator's intent inference capabilities \cite{8206081}. Previous literature identified the conflict for control between the AI agent and the operator as a major challenge in MI \cite{Chiou2019_arXiv, Petousakis2020} and shared control \cite{DeJonge2016}. Typically, conflict for control arises when the operator cannot explicitly communicate their intent. The AI agent attempts to take control of the robot, causing a cycle of the robot and operator overriding each other's commands.

In this work, we aim to infer the operator's intended navigational goal during teleoperation of a remote mobile robot. To address this issue, we consider and build upon the recent recursive Bayesian framework \cite{Jain2018, Jain2019} proposing the following novelties. First, the proposed Bayesian Operator Intent Recognition (BOIR) method enables uncertainty to be taken into account by merging past and present knowledge. The awareness is gathered both implicitly (i.e. by processing sensor and state information) and explicitly by the proposed \textit{action} model (i.e. direct communication of intent by the operator through actions $\sim$ e.g. point-and-click in a displayed map). Second, the method suggests a fusion of multiple observation sources flexibly (i.e. weighting functions with scaling parameters corresponding to the information volume of each evidence relative to the rest). Lastly, the Active Intent Recognition Model (AIRM), which can be useful and adapted to a wide range of applications and interfaces, is proposed to allow the operators to update asynchronously and opportunistically the AI agent beliefs providing explicit information about their intent.

\section{Related Work}
The challenge of understanding the intent of humans is typically met with probabilistic inference and Hidden Markov Models (HMMs) methods. Khokar \textit{et al.} \cite{6907556} used HMMs trained offline to detect human motion intention in the teleoperated system. Other approaches propose the modelling of user's behaviour for intent estimation by analyzing their physiological reactions, such as gaze movement \cite{Aronson2020}.

When the state space is finite, frameworks for both user's intent estimation and planning of supporting actions in assistive robotic applications are proposed using a POMDP formulation \cite{Taha2008}. Other applications include the importance of future action prediction for interactive humanoid robots in manipulation \cite{Wang2013} and trajectory prediction \cite{Best2015}. Predictive neural network model to identify human intended motion based on skeleton-based motion information enabling human-robot co-working is presented in \cite{9035907}. All the above approaches fall into the general scope of planning, action/activity and intent recognition. 

To infer the user's intent, a confidence function is proposed in \cite{Carlson2008} which blends output from two exponential decay models. Similarly, the blending of instantaneous observations (e.g. nearness to the goal, user-generated or robot-generated commands) for evaluating a confidence function is presented in \cite{Gopinath2017, Dragan2013}.

Closer to our research are approaches that utilize Bayes' Theorem to estimate human intent. Finding the intent of robotic wheelchair users in navigational tasks \cite{Huntemann2013} or the intent of assistive manipulation robot users (i.e. which object the user intends to interact with) \cite{Jain2018, Jain2019} is of particular importance. These methods combine present and past information gathered from the user input with a model of the user's driving behaviour and environmental properties.

To the best of the authors' knowledge, most related research links indirect observations (e.g. proximity to goal, joystick signals) to potential intent without allowing for explicit intent inference, e.g. by capturing the user's/operator's intent directly. Additionally, the target applications often assume situations in which the human operator is physically co-located with the robot. In this work, we aim to shift focus and address those gaps by proposing a method for intent recognition that combines both implicit and explicit inference and is applied in a scenario in which the robot is remotely operated on site.

\section{Bayesian Framework, Problem Formulation and Implementation} \label{Bayesian-Framework}
For better intuition, \textit{we formulate the problem} as a disaster response scenario in which a human-robot team is tasked with exploring and inspecting certain points-of-interest (POI). The control method used is teleoperation (i.e. manual joypad control). The map and the POI of the area to be explored are considered known. In a real-world deployment, a map may be available and may be updated or constructed (if absent) with the aid of an unmanned aerial vehicle (UAV) flying over the area of interest and identifying POI (e.g. location of a victim or hazard) that need to be inspected more closely by a mobile robot \cite{Rognon2018}. Those points denote the navigational goals (i.e. set of coordinates) which the operator needs to explore further and hence the robot should be moving towards them. In this paper, we are interested in exploration which is focused to mobile robots moving on the ground plane.

We define a discrete set $\textbf{\textit{G}} = \big\{g^{1}_{t}, g^{2}_{t} ,\dots, g^{N}_{t} \big\}$ containing all known navigational goals $g^i$, where $i$ is an enumerating goal indicator ($i$=1:$N$), $t$ is the current time-step, and $N$ is the number of goals corresponding to POI. The problem we are addressing is to estimate the operator's intent (i.e. which goal the operator is advancing towards) by determining the most probable navigational goal denoted by $g^*\in\textbf{\textit{G}}$. It is considered that the true state of $g\in\textbf{\textit{G}}$ is an unobserved Markov process and the $z\in\textbf{\textit{Z}}$ are the observations of a \textit{Hidden Markov model} (HMM) that gives us information about $g$. To estimate the hidden state $g$ a Bayesian method is proposed (see Algorithm \ref{alg:BOIRalgorithm}). This method is formulated as a mixture of three models. A model that relates observations to the state (\textit{observation} model); a model that indicates the temporal evolution of the state (\textit{transition} model); and a model that captures actions that might occur (\textit{action} model). Once these models are obtained, Bayes' rule is deployed to specify the most probable goal from the total set of goals. 

\begin{equation}\label{final_equation}
\begin{split}
P(g_{t} | z_{1:t}, \alpha_{t}) \propto \underbrace{\prod_{z\in\textbf{\textit{Z}}} P(z_{t} | g_{t})}_\text{observation model} \times \underbrace{\sum_{g_{t-1}} P(g_{t} | g_{t-1})P(g_{t-1})}_\text{transition model} \\ \times \underbrace{{P(g_{t_{0:T-1}} | \alpha_{t_{0:T-1}}})}_\text{action model} 
\end{split}
\end{equation}

The Eq. \ref{final_equation} depicts the core of BOIR method. At each time step $t$, the posterior is calculated, giving us a probability distribution over the $N$ number of goals. Finally, the operator intent $g^\ast$ is inferred as the goal with the highest posterior probability and is acquired by Eq. \ref{max_equation}.

\begin{equation}\label{max_equation}
g^\ast_{t} = \operatorname*{argmax}_{g_{t} \in \textbf{\textit{G}}} P(g_{t} | z_{1:t}, \alpha_{t})
\end{equation}

\subsection{Observation Model}
\label{Observation Model}
For the \textit{observation} model, we consider two geometric observation sources: the angle \footnote[3]{The inverse tangent function $arctan$ is used between robot and goal.} $\phi$ between the robot's pose and each goal and the path length $l$ between the robot and each goal as given by a path planner \footnote[4]{\textit{Global planner} default plug-in in Robot Operating System (ROS) uses the Dijkstra algorithm.}. These observations are considered to be conditionally independent. They are included in the set of observations $\textbf{\textit{Z}} = \big\{\big[\phi^1_{t}, l^1_{t} \big] ,\dots, \big[\phi^{N}_{t}, l^{N}_{t}\big] \big\}$ in pairs. Instead of the commonly used Euclidean distance \cite{Carlson2008, Jain2018}, we exploit path length as a measure for determining the robot's distance from a goal. The latter is advantageous because, in many cases, the robot will have to navigate around obstacles and along winding paths making the Euclidean distance far shorter than the actual path length. The path length is a more informative measurement as it includes environmental context in terms of layout and obstacle distribution. The angle $\phi$ encodes the movement of the robot towards the direction of the goal.

The \textit{observation} model as shown in Eq. \ref{final_equation} is expressed as the product of the likelihoods of observing each evidence. The likelihood models for both observation sources are defined as exponential decay functions of the form $e^{-\frac{z}{w_{z}}}$, where $w_{z}$ is a scaling parameter value assigned to each evidence with $z\in\textbf{\textit{Z}}$. The process of including scaling parameters involves emphasizing the contribution of particular evidence over others to the final result. Simply put, they define how each observation source will be taken into account relative to the others. In addition to the mathematical convenience, the parameters have an intuitive interpretation that can be easily adapted to reflect how similar types of evidence (i.e. numerical geometric measurements) are assumed to carry less or more information about the operator's intent. The scaling parameters satisfy the condition of summing to one, and to have a meaningful impact, the input values from both observation sources are \say{{min-max}} normalized to the range $[0,1]$, before joining the data set \textbf{\textit{Z}}.

\subsection{Transition Model}
\label{Transition Model}
This model allows us to provide information about the predicted (i.e. current) state given the immediately previous one. The probability distribution of the predicted state is the sum of the products of the probability distribution associated with the goal transition from the $t-1$ time step to the $t$ and the probability distribution associated with the previous state, over all possible goals $g_{t-1}$. The first term of the \textit{transition} model in Eq. \ref{final_equation} represents the conditional probability \cite{Baker2007, Jain2018, Huntemann2013} distribution over changing to the goal $g$ at time $t$ given that the goal was $g$ at time $t-1$ (Eq. \ref{transition_equation}).
\begin{equation}\label{transition_equation}
       P(g_{t}=g^i | g_{t-1}=g^j) = 
        \begin{cases}
            1-\Delta & \text{if $i=j$} \\
            \frac{\Delta}{N-1} & \text{otherwise}
        \end{cases}
\end{equation}


The second term denotes the prior belief over all goals, which is recursively updated as the posterior calculated in the algorithm's previous iteration. At the beginning of each experimental run, this prior belief is uniformly distributed over $N$ hypotheses (i.e. $N$ number of goals). 

\begin{algorithm} 
\caption{Bayesian Operator Intent Recognition (BOIR)}
\label{alg:BOIRalgorithm}
\begin{algorithmic}[]
\Require \State known environment, $goals \in G$, time horizon $(T)$, activated Belief $(\lambda)$, threshold Belief $(tB)$, $w_{\phi}$, $w_{l}$, $\Delta$
\end{algorithmic}
\begin{algorithmic}[1]
\Procedure{Bayesian Recognition}{} 
    \State Initialize uniform $prior$ distribution: $P(g_{t=0})$
    \State Define $Posterior_{t=0}(g) = P(g_{t=0})$
    
    \State $action (\alpha) \sim \text{AIRM}$
    \While{(termination conditions not met)}
    \State Make observations $z \in Z$ (angle \& path)
    \If{$action (\alpha) = True$}
    \State Define AIRM $\leftarrow \eqref{AIRM}$ \Comment{$T$ starts}
    \Else
    \State \textbf{\textit{pass}} \Comment{no effect}
    \EndIf
    \For{$\forall g\in\textbf{\textit{G}}$}
    \State Compute $Posterior \leftarrow \eqref{final_equation}$
    \EndFor
    \State Normalize result
    \State Update operator intent $g^\ast_{t} \leftarrow \eqref{max_equation}$
    \State Set $prior_{t+1} \leftarrow Posterior_{t}$
    \EndWhile  
\EndProcedure

\end{algorithmic}
\end{algorithm}

\subsection{Action Model - AIRM}
\label{Action Model}
The last model shown in Eq. \ref{final_equation} referring to the proposed AIRM (Eq. \ref{AIRM}), indicates the probability of a goal given the explicit information that has been provided by the operator. The AIRM allows BOIR method to receive asynchronously and on-demand explicit knowledge about the operator's intent once and if it is available. We incorporate the operator's explicit communication of intent via an appropriate interface into the Bayesian probabilistic model. This knowledge is registered through a point-and-click input in the Graphical User Interface, 
which corresponds to a waypoint to the intended goal. This can be useful in various circumstances, e.g. when a sudden change of operator's intent might occur, when multiple or overlapping goals exist, or cases in which human intent is not obvious and difficult to be decided accurately, since human behavior is characterized by bounded rationality \cite{10.1257/000282803322655392}. In practice, the AIRM can be trivially adapted to a number of different configurations to enable a wide variety of other interfaces to be presented, e.g. multi-modal intent communication, such as voice commands via natural language and/or via eye/gaze tracking. The AIRM is defined in the following way:
\begin{equation}\label{AIRM}
        P(g^y_{t_{0:T-1}} | \alpha_{t_{0:T-1}}) = 
       \begin{cases}
            -R \cdot t + \lambda &, \text{$t$=$0$:$T$-$1$}
             \\
            1 &, \text{otherwise}
        \end{cases}
\end{equation}
where,
\begin{equation}\label{rate_equation}
R = \frac{\lambda - tB}{T}
\end{equation}

We let $\lambda$ be the activated Belief parameter that defines the probability of the selected goal $y$ at the initial time step of the activation. After the intent is communicated to the AIRM, the probability of the selected goal is undergoing a linear decay at a rate $R$ (see Eq. \ref{rate_equation}). We let $tB$ be the threshold Belief marking the probability of the selected goal, whose value can never fall below, during the activated time horizon $T$. Subsequently, the probability of the selected goal $y$ will be higher, whereas the probability of the remaining goals will be equally distributed. This activation at $t=0$ also affects the second term of the \textit{transition} model (i.e. prior belief) by simply labelling it with the same initial probability values for each goal.

\begin{figure}
	\centerline{\subfigure[]{\includegraphics[width=0.48\columnwidth]{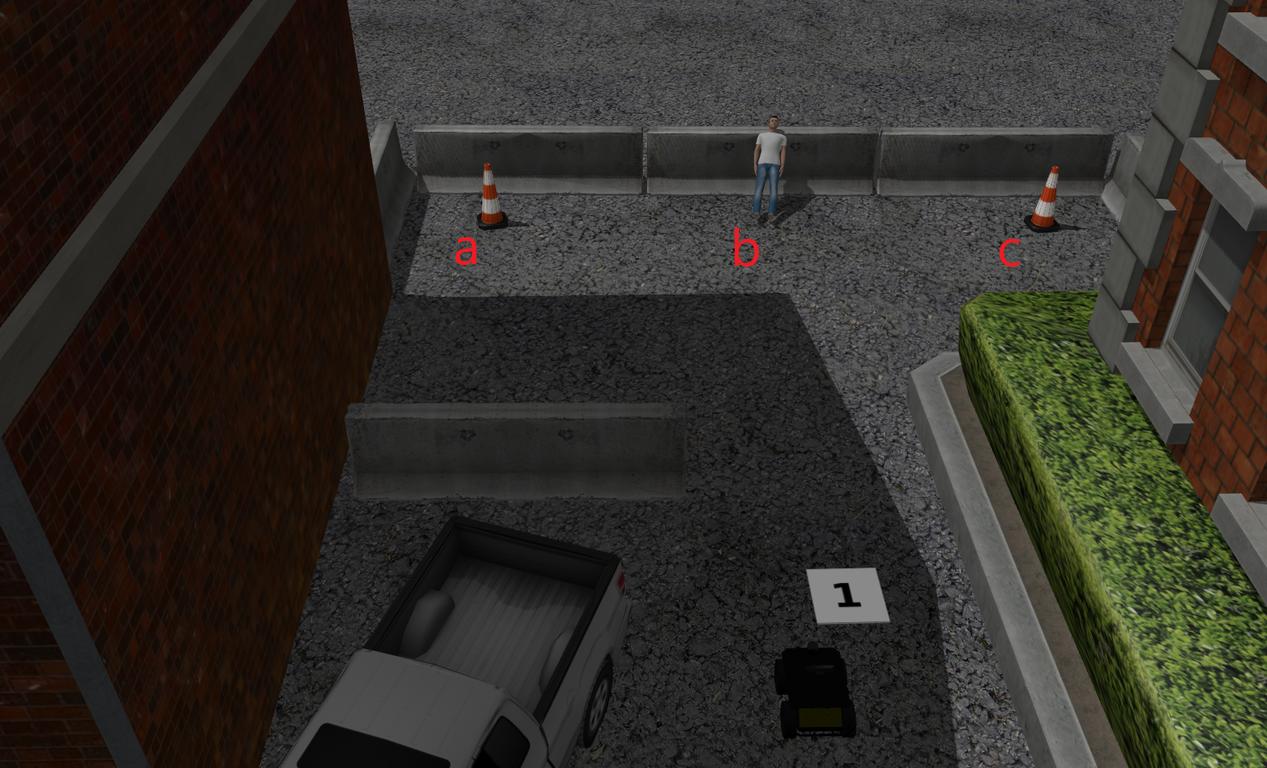}
			\label{fig:area1}}
		\hfil
		\subfigure[]{\includegraphics[width=0.48\columnwidth]{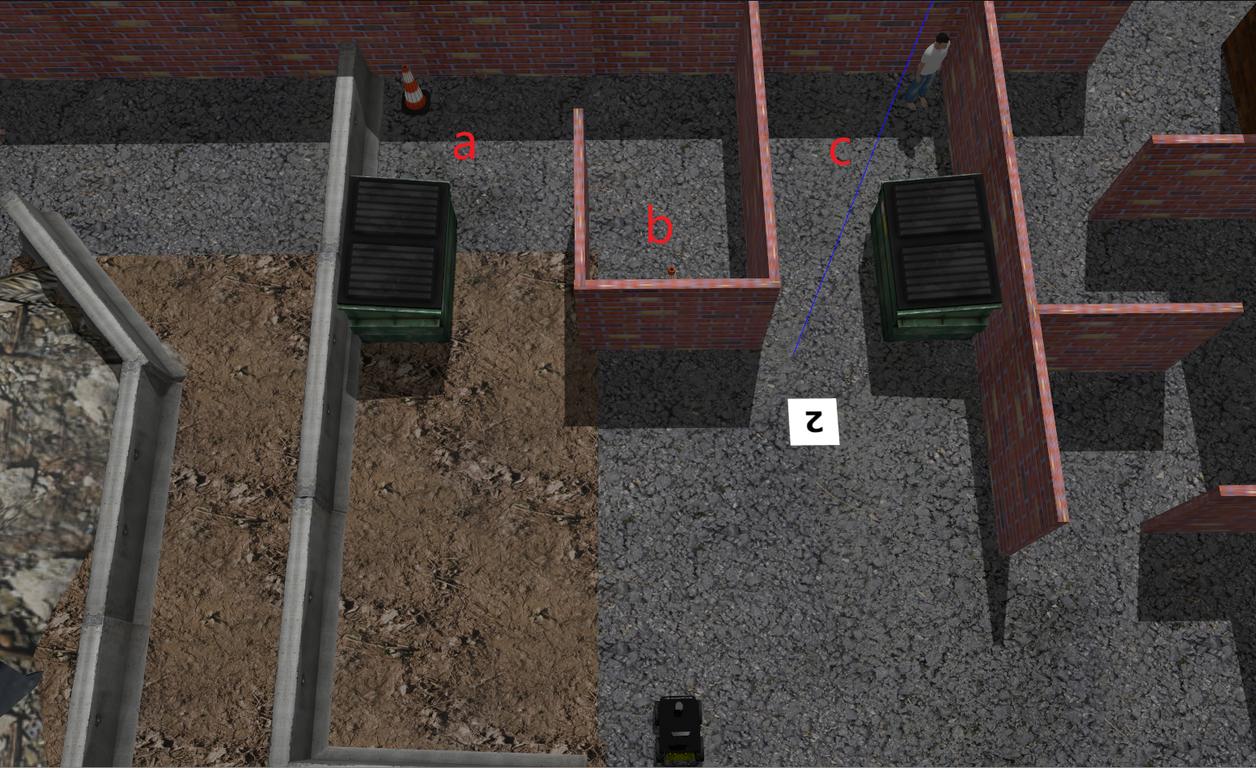}
			\label{fig:area2}}}
		\hfil
		\subfigure[]{\includegraphics[width=0.48\columnwidth]{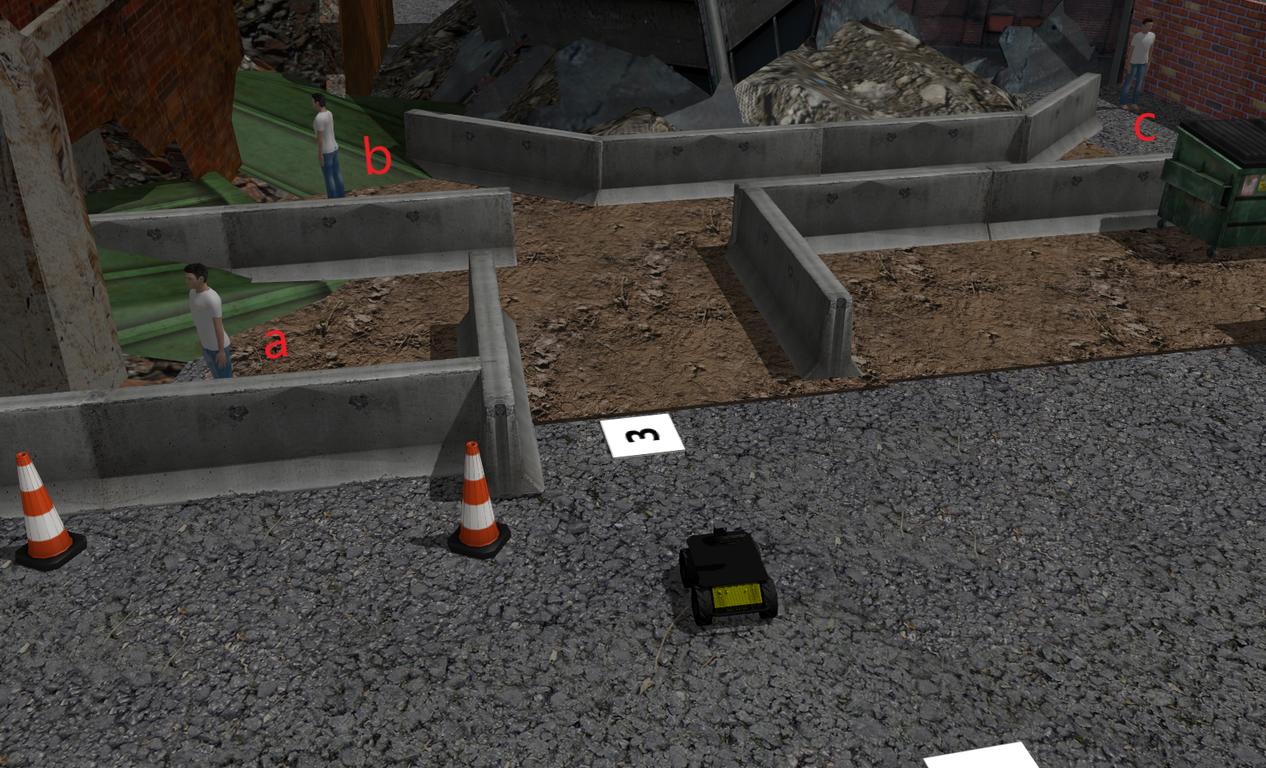}
			\label{fig:area3}}
		\hfil
		\subfigure[]{\includegraphics[width=0.475\columnwidth]{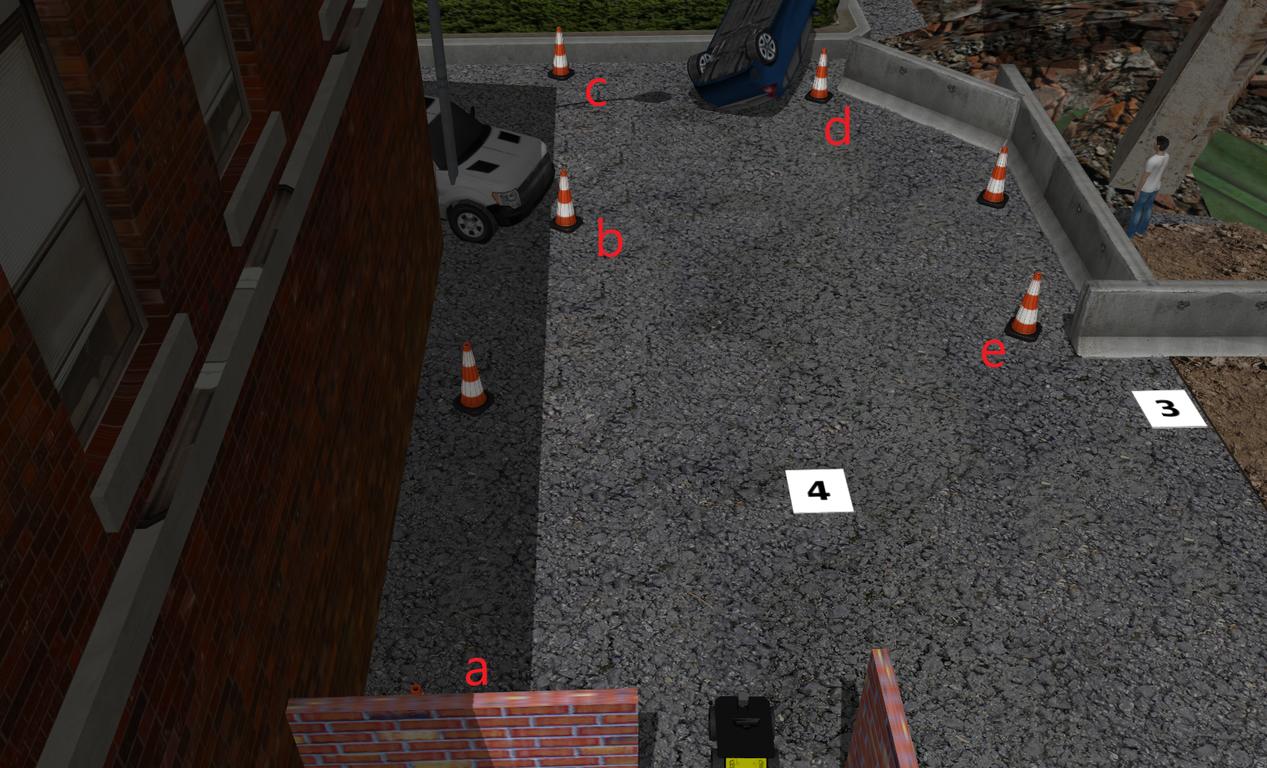}
			\label{fig:area4}}
	\caption{\textbf{\ref{fig:area1}}-\textbf{\ref{fig:area4}:} The evaluation scenarios. The numbers denote the scenarios; the lower-case letters denote the points-of-interest for exploration (i.e. the set of possible goals); the point in which the husky robot stands refers to the starting point of each scenario; the humans denote the intended goals (with the exception of scenario 4).}
	\label{fig:scenarios}
\end{figure}

\section{Experimental Study}
An experimental evaluation of the proposed approach of Section \ref{Bayesian-Framework}, inspired by disaster response and remote inspection scenarios, was conducted to evaluate the functionality of the proposed BOIR method based on the following aspects: a) the accuracy and uncertainty of predicting the intended goal; b) the effect of complex map layout on prediction (e.g. increased number of goals, sequential goals, complex manoeuvres, obstacle layout profile). 

\subsection{Apparatus \& Software}
Several experimental scenarios were conducted on a realistic robotic simulator named Gazebo. The simulation was used to avoid introducing confounding factors from a real-world robot trial and improve the experiment's repeatability. The software was developed in Robot Operating System (ROS) \footnote[5]{The code for the experiments and BOIR is provided under MIT license in the Extreme Robotics Lab GitHub repository: \url{https://github.com/uob-erl/operator_intent_packages}} and the simulated robot was a Husky model with a camera to provide video feed and a laser scanner for localization and object detection. The robot was teleoperated via an Operator Control Unit.

\subsection{Experimental Procedure \& Scenarios}
\label{Experimental_Procedure_Scenarios}
A total of 4 participants took part in the study. They were 4 males with a mean age of 28.5 years. All of them had extensive previous experience in operating similar robotic systems and were proficient or experts in using the OCU and the GUI.

The operators were tasked with navigating between different POI, i.e. goals that need to be inspected more closely by a mobile robot. The POI, which are denoted by lower-case letters, are within multiple areas with the latter suggesting a different scenario (see Fig. \ref{fig:scenarios}). Each operator performed five trials on each of the scenarios, resulting in a total of 20 trials. The four scenarios are described in depth below.


\textbf{Scenario 1} represents the most simple case where the operators were tasked with moving from the starting point to goal \say{b} and midway change their intent to \say{a} (see Fig. \ref{fig:area1}).

\textbf{Scenario 2} features a more complex map layout with the different goals being obstructed behind obstacles forcing the operator to maneuver around them to reach the intended goal. In all trials the operators' intent was to reach goal \say{c} (see Fig. \ref{fig:area2}). 

\textbf{Scenario 3} serves to assess the system's performance in a sequential goal setup. It consisted of three sub-areas containing POI (i.e. goals) that the operators had to navigate in sequence from the starting pose to \say{a}, \say{b}, and finish in \say{c} (see Fig. \ref{fig:area3}). Each change of goal in this sequence marks a change in the operator's intent.

\textbf{Scenario 4} entails that the proposed method can be adapted to a bigger and different than the local context by introducing multiple potential goals. The operators were tasked with navigating between two POI, randomly chosen in the beginning of each trial using random number generation. Similarly, the switch from one goal to another implies a change in the operator's intent. A total of five goals where dispersed in the navigation region (see Fig. \ref{fig:area4}). This was done to ensure variance and simulate the cases that the operator might explore in a less efficient and more varied way, e.g. might return to the area's entrance or pass by an already explored POI.

As can be seen, in all scenarios, except the second one, the operator had to give up on the initial goal before or after reaching it and switch to another goal. The aim was to evaluate how the proposed method reacts both to \say{single intent} prediction (scenario 2) and \say{change of intent} prediction (scenarios 1, 3, 4).

\subsection{Implementation - Model Parameters}
In our default setting, we consider the following model parameters: $n$ is known (3 or 5; depending on scenario (see Section \ref{Experimental_Procedure_Scenarios})), $w_{\phi}=0.6$, $w_{l}=0.4$, $\Delta=0.2$, $\lambda=0.95$, $tB=0.35$, $T=10s$. In the implementation used in this paper, the parameters were empirically calculated. Parameters' optimization is beyond the scope of this paper.

\subsection{Comparison \& Metrics}
The performance of the proposed method is compared to two methods from the literature considering the following metrics: a) the percentage of correct predictions over the span of each trial (i.e. accuracy) and b) the cross-entropy (log-loss) function \cite{Goodfellow-et-al-2016} for assessing the uncertainty of a prediction by defining the divergence between the predicted and the actual value. Specifically, we compare our BOIR to the Recursive Bayesian Intent Inference (RBII-1, observation scheme one) method \cite{Jain2018,Jain2019} whose structure is considered a state-of-the-art Bayesian intent estimation framework in manipulation tasks and it is shown to outperform other methods. In particular, it considers the evolution of the state similar to ours and the Euclidean distance between the robotic arm and the goal. The second algorithm \cite{Carlson2008} we compare to is representative of commonly used intent prediction methods in which a confidence function defined as the product of two exponential models is used. We will refer to this method as Exponential Confidence Function (ECF). The method proposed here should not be seen as competing with such schemes, but rather as complementing them.

The second step of the experimental study involves running and testing several trials in scenarios 3 and 4 where the operators use the AIRM functionality denoting their \say{change of intent} explicitly. We test and evaluate the proposed BOIR-AIRM and compare it to BOIR, RBII-1 and ECF. 

\begin{figure}
	\centering
	\includegraphics[width=0.95\columnwidth]{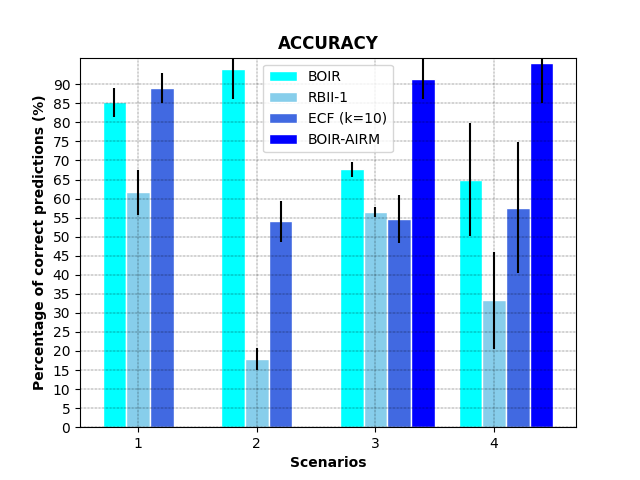}
	\caption{Accuracy (i.e. the percentage of correct predictions from the total predictions made) performance comparison across each method in every task scenario. Plot shows the means and standard deviations.}
	\label{fig:acc_plot}
\end{figure}

\begin{figure}
	\centering
	\includegraphics[width=0.95\columnwidth]{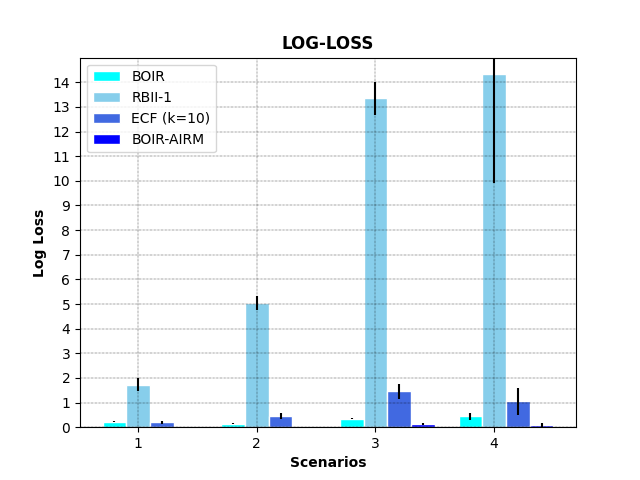}
	\caption{Log-loss performance comparison across each method. The log-loss metric measures the uncertainty of the predictions by penalizing those predictions that are confident and wrong. A perfect model would have a log-loss of 0.} 
	\label{fig:logloss_plot}
\end{figure}


\section{Results}
The experimental results and descriptive statistics are summarized in Table \ref{table:results} and visualized in Fig. \ref{fig:acc_plot} and \ref{fig:logloss_plot}. The results of the pairwise comparisons are shown in Table \ref{table:statistical_analysis}. For the data that conformed to a normal distribution, paired sample t-test statistic was used while for the rest the Wilcoxon signed-rank test was reported. Bonferroni post-hoc correction (significant at the $0.05$ level) was applied to the p-values presented (i.e. the correction $c$ matching the number of pairwise comparisons, $c = 3$ for scenarios 1 and 2 and $c = 6$ for the rest). 

\begin{table*}[t]
    \caption{Descriptive statistics for accuracy \& Log-Loss}
	\centering
	
	    \begin{tabular}{lllll}
		\hline
		\textbf{Scenarios \& Methods}                         
		& \textbf{effect size}&      \textbf{descriptive statistics} 
		
		& \textbf{effect size}
		&      \textbf{descriptive statistics}\\
		& & \textbf{(accuracy)} & &\textbf{(log-loss)} \\
		\hline
		\begin{tabular}[c]{@{}l@{}}  \textit{Scenario 1} \\ BOIR \\  RBII-1 \\ ECF \end{tabular}
		&
		\begin{tabular}[c]{@{}l@{}}\\$\eta^2 = .878$\end{tabular}
		& 
		\begin{tabular}[c]{@{}l@{}}\\$M = 85.30$ $\%$, $SD = 3.83$ \\ $M = 61.60$ $\%$, $SD = 5.88$ \\ $M = 89.10$ $\%$, $SD = 3.96$\end{tabular}
		&
		\begin{tabular}[c]{@{}l@{}}\\$\eta^2 = .953$\end{tabular}
		&
		\begin{tabular}[c]{@{}l@{}}\\$M = 0.222$, $SD = 0.03$\\ $M = 1.73$, $SD = 0.28$\\ $M = 0.194$, $SD = 0.05$\end{tabular}
		\\  \hline
		\begin{tabular}[c]{@{}l@{}}\textit{Scenario 2} \\ BOIR \\ RBII-1 \\ ECF \end{tabular}
		&
		\begin{tabular}[c]{@{}l@{}}\\$\eta^2 = .968$\end{tabular}
		&  
		\begin{tabular}[c]{@{}l@{}}\\$M = 94.00$ $\%$, $SD = 7.85$ \\ $M = 17.90$ $\%$, $SD = 2.99$ \\ $M = 54.00$ $\%$, $SD = 5.42$\end{tabular}
		&
		\begin{tabular}[c]{@{}l@{}}\\$\eta^2 = .994$\end{tabular}
		&
		\begin{tabular}[c]{@{}l@{}}\\$M = 0.133$, $SD = 0.023$\\ $M = 5.034$, $SD = 0.28$\\ $M = 0.464$, $SD = 0.124$\end{tabular}
		\\\hline
		\begin{tabular}[c]{@{}l@{}}\textit{Scenario 3} \\ BOIR \\ BOIR-AIRM \\ RBII-1 \\ ECF \\ \end{tabular}
		&
		\begin{tabular}[c]{@{}l@{}}\\$\eta^2 = .925$\end{tabular}& 
		\begin{tabular}[c]{@{}l@{}}\\$M = 67.65$ $\%$, $SD = 2.03$ \\ $M = 91.45$ $\%$, $SD = 5.30$\\ $M = 56.50$ $\%$, $SD = 1.32$ \\ $M = 54.60$ $\%$, $SD = 6.26$\end{tabular}
		&
		\begin{tabular}[c]{@{}l@{}}\\$\eta^2 = .996$\end{tabular}
		&
		\begin{tabular}[c]{@{}l@{}}\\$M = 0.35$, $SD = 0.02$\\ $M = 0.122$, $SD = 0.03$ \\ $M = 13.35$, $SD = 0.68$\\ $M = 1.46$, $SD = 0.3$\end{tabular}
		\\\hline
		\begin{tabular}[c]{@{}l@{}}\textit{Scenario 4} \\ BOIR \\ BOIR-AIRM \\ RBII-1 \\ ECF \end{tabular}             
		&
		\begin{tabular}[c]{@{}l@{}}\\$\eta^2 = .726$\end{tabular} &
		\begin{tabular}[c]{@{}l@{}}\\$M = 65.00$ $\%$, $SD = 14.72$ \\ $M = 95.55$ $\%$, $SD = 10.56$\\ $M = 33.30$ $\%$, $SD = 12.79$ \\ $M = 57.60$ $\%$, $SD = 17.17$\end{tabular}
		&
		\begin{tabular}[c]{@{}l@{}}\\$\eta^2 = .883$\end{tabular}
		&
		\begin{tabular}[c]{@{}l@{}}\\$M = 0.44$, $SD = 0.15$\\ $M = 0.08$, $SD = 0.1$ \\$M = 14.35$, $SD = 4.45$\\$M = 1.05$, $SD = 0.55$\end{tabular}
		\\ \hline
	    \end{tabular}
	\label{table:results}
\end{table*}

\begin{table*}[]
    \caption{Pairwise comparisons with Bonferroni corrections applied to the p-values.}
	\centering
	
	    \begin{tabular}{lllll}
		\hline
		\textbf{Scenarios \& Methods}   
	    & \textbf{statistic} &\textbf{corrected p-value}& \textbf{statistic} &\textbf{corrected p-value} \\
	    
	    & & \textbf{(accuracy)} & &\textbf{(log-loss)}
		
		\\ \hline
		\begin{tabular}[c]{@{}l@{}}  \textit{Scenario 1} \\ BOIR vs RBII-1 \\ BOIR vs ECF \\ ECF vs RBII-1 \end{tabular}
		&
		\begin{tabular}[c]{@{}l@{}}\\$t(19) = 20.25$ \\ $t(19) = -10.78$ \\ $t(19) = 20.60$ \end{tabular}
		& 
		\begin{tabular}[c]{@{}l@{}}\\$p < .001$ \\ $p < .001$ \\ $p < .001$\end{tabular}
		&
		\begin{tabular}[c]{@{}l@{}}\\$t(19) = -25.56$ \\ $t(19) = 3.58$ \\ $t(19) = -28.12$ \end{tabular}
		& 
		\begin{tabular}[c]{@{}l@{}}\\$p < .001$ \\ $p < .01$ \\ $p < .001$\end{tabular}
		\\  \hline
		\begin{tabular}[c]{@{}l@{}}\textit{Scenario 2} \\ BOIR vs RBII-1 \\ BOIR vs ECF \\ ECF vs RBII-1 \end{tabular}
		&
		\begin{tabular}[c]{@{}l@{}}\\$Z = -3.92$ \\ $Z = -3.93$ \\ $Z = -3.93$\end{tabular}
		&  
		\begin{tabular}[c]{@{}l@{}}\\$p < .001$ \\ $p < .001$ \\ $p < .001$\end{tabular}
		&
		\begin{tabular}[c]{@{}l@{}}\\$Z = -3.92$ \\ $Z = -3.92$ \\ $Z = -3.92$\end{tabular}
		&  
		\begin{tabular}[c]{@{}l@{}}\\$p < .001$ \\ $p < .001$ \\ $p < .001$\end{tabular}
		\\\hline
		\begin{tabular}[c]{@{}l@{}}\textit{Scenario 3} \\ BOIR vs RBII-1 \\ BOIR vs ECF \\ ECF vs RBII-1 \\ BOIR-AIRM vs BOIR \\ BOIR-AIRM vs RBII-1 \\ BOIR-AIRM vs ECF \\ \end{tabular}
		&
		\begin{tabular}[c]{@{}l@{}}\\$t(19) = 27.48$ \\ $t(19) = 10.21$ \\ $t(19) = 10.21$ \\ $Z = -3.93$ \\ $Z = -3.92$ \\ $Z = -3.92$\end{tabular}
		& 
		\begin{tabular}[c]{@{}l@{}}\\$p < .001$ \\ $p < .001$ \\ $p = .95$ \\ $p < .001$ \\ $p < .001$ \\ $p < .001$\end{tabular}
		& 
		\begin{tabular}[c]{@{}l@{}}\\$t(19) = -87.64$ \\ $t(19) = -17.41$ \\ $t(19) = -87.08$ \\ $Z = -3.93$ \\ $Z = -3.92$ \\ $Z = -3.92$\end{tabular}
		& 
		\begin{tabular}[c]{@{}l@{}}\\$p < .001$ \\ $p < .001$ \\ $p < .001$ \\ $p < .001$ \\ $p < .001$ \\ $p < .001$\end{tabular}
		\\\hline
		\begin{tabular}[c]{@{}l@{}}\textit{Scenario 4} \\ BOIR vs RBII-1 \\ BOIR vs ECF \\ ECF vs RBII-1 \\ BOIR-AIRM vs BOIR \\ BOIR-AIRM vs RBII-1 \\ BOIR-AIRM vs ECF \\ \end{tabular}     
		&
		\begin{tabular}[c]{@{}l@{}}\\$t(19) = 16.97$ \\ $t(19) = 3.19$ \\ $t(19) = 8.53$ \\ $Z = -3.92$ \\ $Z = -3.93$ \\ $Z = -3.92$\end{tabular}
		& 
		\begin{tabular}[c]{@{}l@{}}\\$p < .001$ \\ $p < .05$ \\ $p < .001$ \\ $p < .001$ \\ $p < .001$ \\ $p < .001$\end{tabular}
		&
		\begin{tabular}[c]{@{}l@{}}\\$Z = -3.92$ \\ $Z = -3.92$ \\ $Z = -3.92$ \\ $Z = -3.92$ \\ $Z = -3.92$ \\ $Z = -3.92$\end{tabular}
		& 
		\begin{tabular}[c]{@{}l@{}}\\$p < .001$ \\ $p < .001$ \\ $p < .001$ \\ $p < .001$ \\ $p < .001$ \\ $p < .001$\end{tabular}
		\\ \hline
	    \end{tabular}
	\label{table:statistical_analysis}
\end{table*}

The results show that BOIR predicts the correct goal more often (i.e. higher accuracy) when compared to the other methods, except for scenario 1 in which ECF outperforms BOIR in terms of accuracy by $3\%$. In scenarios 2, 3 and 4 BOIR yields the highest accuracy for both \say{single intent} and \say{change of intent} prediction. Similarly, log-loss results suggest that in scenario 1 ECF maintains a better performance with less uncertainty, however, in scenarios 2, 3 and 4, BOIR outperforms both ECF and RBII-1. The latter method demonstrates significantly higher uncertainty. A further comparison shows that in scenarios 3 and 4 BOIR's overall performance is significantly improved when the operators use the AIRM to communicate their intent. 

\section{Discussion}
Despite evaluation scenarios becoming increasingly complex (e.g. increased number of goals, sequential goals, complex map layouts), the proposed method managed to overcome these hindrances and outperform the other methods. One explanation is that the incorporation of the path length (hence the incorporation of map layout knowledge), over the Euclidean distance gives BOIR an advantage. Compared to the other methods, BOIR is not biased towards the closer goal but instead it is jointly formed using the shortest path and the angle between the robot and the goal. While ECF and RBII-1 methods only predict the correct goal as the robot gets much closer to it, it was observed that BOIR can identify the goal much earlier. This is especially true in scenario 2 (see Fig. \ref{fig:area2}) in which the other methods recognize the goal \say{b} as the most probable, while the correct is the goal \say{c}.

The results show that in scenarios 3 and 4, there is a drop in the performance of BOIR compared to other scenarios. This degradation could be attributed to the fact that the operators required to visit an increased number of goals in a complex sequence. In scenario 3 (see Fig. \ref{fig:area3}), the initial pose of the robot indicates that the operator intends to move towards goal \say{b}, however, the initial goal needed to be inspected was \say{a}. This could be explained by the fact that path lengths for both goals \say{a} and \say{b} are approximately equal while the angle between the latter and the robot is much smaller. Consequently, the algorithm cannot capture this intent, leading to some wrong predictions. Similar situations were observed in scenario 4, suggesting mainly that in cases where multiple goals are \say{aligned}, the intent can become ambiguous making several interpretations plausible. This problem, as results showed, could be addressed with the AIRM functionality, which gives a stronger prior belief to the goal chosen explicitly.

Furthermore, the results provide support for the proposed observation model and the way it treats the information gathered from the sensors. The unity-based normalization to the range [0, 1] is to ensure that the observation values measured on different scales will be adjusted to a notionally common scale. The weighting functions then will handle each independent normalized value equally and decide what their influence on the result would be based on the scaling parameters.

Since the target domain of this paper is time and safety critical applications, the emphasis on robustness and stability of predictions is of primary importance \cite{Delmerico2019_SAR_JFR}. Hence, it is crucial that BOIR, based on the evidence showing high accuracy and low uncertainty, avoids situations in which the continuous alternation among the potential goals becomes frequent as it may disrupt the system's efficient use.


In the current implementation, BOIR's parameters are heuristically chosen. In future work, optimization techniques can be used to learn the optimal values, especially those that refer to the observation model. Moreover, we are assuming that the set of goals (POI) is \say{a priori} known, meaning neither the human nor the robot are tasked with identifying new goals during the exploration. Extending the algorithm to dynamically cope with a changing set of goals can be used alongside AI recognition algorithms (e.g. via computer vision, infrared sensors) to analyze the scene to determine the location of new goals such as trapped victims, leaking pipes, radiation sources, or other hazards.

\section{Conclusion}
In this paper, we propose and empirically validate a probabilistic method named BOIR capable of recognizing human operator intended navigational goal. Effectiveness of the proposed method is demonstrated through experimental evaluation in navigation scenarios and compared to the state-of-the-art algorithms in the field. Various map layouts, including several POI, are incorporated which the operators are tasked with inspecting by controlling a remotely operated mobile robot. A variety of new contributions are made, including the incorporation of more informative observation sources that reflect the layout of an area; the benefit of combining multiple observation sources in a flexible and weighting way; and the proposal of the AIRM model that enables operators to provide their intent explicitly, asynchronously, and opportunistically. In future work, by extending our focus on variable autonomy human-robot teams such as in Mixed-Initiative systems, BOIR can be used to dynamically inform the autonomous and/or semi-autonomous policies enabling such robotic systems to actively and efficiently assist human operators.


\addtolength{\textheight}{-9.33cm}   


\bibliography{thebibliography}
\bibliographystyle{IEEEtran}

\end{document}